\documentclass[conference]{IEEEtran}

\usepackage[utf8]{inputenc}
\usepackage{multirow}
\usepackage[inline]{enumitem}
\usepackage{xcolor}
\usepackage{booktabs}
\usepackage[pdfauthor={Alexander Hartl, Maximilian Bachl, Joachim Fabini, Tanja Zseby},
pdftitle={Explainability and Adversarial Robustness for RNNs}]{hyperref}
\usepackage{amsmath}
\usepackage{amssymb}
\usepackage{graphicx}
\usepackage[numbers]{natbib}
\usepackage{subfig}
\usepackage{tikz}

\usepackage[nomain, toc, acronym]{glossaries}
\glsdisablehyper

\begin{document}

\title{Explainability and Adversarial Robustness for RNNs}

\author{\IEEEauthorblockN{Alexander Hartl,
Maximilian Bachl, Joachim Fabini, Tanja Zseby}
\IEEEauthorblockA{Technische Universität Wien\\
firstname.lastname@tuwien.ac.at}}




\newcommand\copyrighttext{%
  \footnotesize \textcopyright 2020 IEEE. Personal use of this material is permitted.
  Permission from IEEE must be obtained for all other uses, in any current or future
  media, including reprinting/republishing this material for advertising or promotional
  purposes, creating new collective works, for resale or redistribution to servers or
  lists, or reuse of any copyrighted component of this work in other works.}
\newcommand\copyrightnotice{%
\begin{tikzpicture}[remember picture,overlay]
\node[anchor=south,yshift=10pt] at (current page.south) {\fbox{\parbox{\dimexpr\textwidth-\fboxsep-\fboxrule\relax}{\copyrighttext}}};
\end{tikzpicture}%
}

\maketitle%
\copyrightnotice

\newacronym{ml}{ML}{Machine Learning}
\newacronym{dl}{DL}{Deep Learning}
\newacronym{aml}{AML}{Adversarial Machine Learning}
\newacronym{ids}{IDS}{Intrusion Detection System}
\newacronym{rnn}{RNN}{Recurrent Neural Network}
\newacronym{fgsm}{FGSM}{Fast Gradient Sign Method}
\newacronym{cw}{CW}{Carlini-Wagner method}
\newacronym{pgd}{PGD}{Projected Gradient Descent}
\newacronym{pdp}{PDP}{Partial Dependence Plot}
\newacronym{ars}{ARS}{Adversarial Robustness Score}
\newacronym{ttl}{TTL}{Time-to-Live}
\newacronym{dos}{DoS}{Denial-of-Service}
\newacronym{iat}{IAT}{Interarrival time}

\begin{abstract}

\glspl{rnn} yield attractive properties for 
constructing \glspl{ids} for network data.
With the rise of ubiquitous \gls{ml} systems, malicious actors have been catching up quickly to find new ways to exploit \gls{ml} vulnerabilities for profit. Recently developed adversarial \gls{ml} techniques focus on computer vision and their applicability to network traffic is not straightforward: Network packets expose fewer features than an image, are sequential and impose several constraints on their features. 

We show that despite these completely different characteristics, adversarial samples can be  generated reliably for \glspl{rnn}.
To understand a classifier's potential for misclassification, we extend existing explainability techniques and propose new ones, suitable particularly 
for sequential data. Applying them shows that already the first packets of a communication flow are of crucial importance and are likely to be targeted by attackers. Feature importance methods show that even relatively unimportant features can be effectively abused to generate adversarial samples. We thus introduce the concept of \textit{feature sensitivity} which quantifies how much potential a feature has to cause misclassification. 

Since traditional evaluation metrics such as accuracy are not sufficient for quantifying the adversarial threat, we propose the \gls{ars} for comparing \glspl{ids} 
and show that an adversarial training procedure can significantly and successfully reduce the attack surface.

\end{abstract}


\maketitle

\section{Introduction}

There is a significant body of scientific work focusing on the detection of unwanted behavior in networks. In the past, a viable way of performing intrusion detection was to inspect the content of packets themselves and detect if a packet delivers potentially harmful content. More recently, with the increasing deployment of encryption, 
the focus now lies on features that are always available to network monitoring equipment like packet sizes, protocol flags or port numbers when encrypting above the transport layer.

Network communication is usually aggregated into \textit{flows}, which are commonly defined as a sequence of packets sharing certain properties. 
When analyzing flows, not only the aforementioned features are available but also features related to the timing of the individual packets. 
Various approaches have been proposed to extract features from flows and then perform anomaly detection with the extracted flows \cite{meghdouri_analysis_2018}.
While these approaches frequently work well, it is problematic that the whole flow has to be received first and only afterwards anomaly detection is applied, revealing attack flows. Thus, we design a network \gls{ids} that operates on a per-packet basis and decides if a packet is anomalous based on features that are available even for traffic that is encrypted above the transport layer, like for example TLS or QUIC.
At the same time, an \gls{rnn}-based \gls{ids} has the benefit of providing any available information to the classifier while avoiding tedious feature engineering procedures, which derive statistical measures from the sequence of packet features.
We show that our system has similar performance to other flow-based anomaly detection systems but can detect anomalies \textit{before} the flow terminates.

However, for practical use, high detection accuracy is not enough. With the recent rise of interest in \gls{aml} techniques, also \gls{aml} for \glspl{ids} has been investigated. For example, \cite{bachl_walling_2019} investigates remedies for poisoning attacks on \glspl{ids} and \cite{hashemi_towards_2019} investigate adversarial robustness of common network \glspl{ids}.
In this research, we investigate whether adversarial samples, i.e. minimally different attack flows which are classified as benign, can be found for our \gls{rnn}-based model. This is not a trivially answerable question since adversarial samples have mostly been analyzed in the context of computer vision. Our scenario significantly deviates from computer vision because \begin{enumerate*}
\item the number of features is significantly smaller and
\item only certain features can be manipulated if the flow should remain valid. 
\end{enumerate*}

Surprisingly, we can confirm that adversarial samples can be found even when considering these tight constraints. 
Thus, we argue that traditional performance metrics like, e.g., accuracy are not sufficient for evaluating security-sensitive \gls{ml} systems.
Despite the well-known threat of adversarial samples, we find that related literature lacks metrics for adversarial robustness and, in particular, no measure has been proposed for quantifying the robustness of \gls{ml}-based \glspl{ids}.

Due to the threat of adversarial attacks, but also as a basic requirement for social acceptance of an \gls{ml}-based system, a further crucial requirement for \gls{ml}-based \glspl{ids} is that the classifier's decisions are interpretable by humans. This recently stirred up increased interest in explainability methods. Also in this case, common methods are designed to work with images or tabular data, but not with sequences. 

In this paper, we attempt to provide a comprehensive discussion of these security-related topics in the context of \glspl{rnn}. Our main contributions are:
\begin{itemize}[leftmargin=0.5cm]
\item We analyze several methods for generating adversarial samples and show that adversarial samples can be generated efficiently for an \gls{rnn}-based classifier.
\item Based on common robustness notions of related works, we propose the \acrfull{ars} as a new performance score for \glspl{ids}, which captures the notion of how easily an adversary can generate adversarial samples. We demonstrate that the \gls{ars} can be computed efficiently.
\item We review methods for evaluating which features have a significant impact on the classifier's prediction, both picking up methods that have been proposed in the literature, extending them for \glspl{rnn}, and devising new methods. 
Astonishingly, feature importance methods reveal that
features, which are manipulated for successful adversarial flows,
are not even particularly important for the \gls{rnn}'s classification outcome. 
Thus, we propose \textit{feature sensitivity} methods, which show how prone a feature is to cause misclassification. 

\item Going further, we investigate which packets have a significant contribution to the classifier's decision and which values of these features lead to classification as attack. Hence, we extend existing explainability methods such as \glspl{pdp} \cite{friedman_greedy_2001} for sequential data. 

\item
Based on the insights gained by the feature importance and explainability methods, we finally propose two defense methods.
First, by simply leaving out manipulable features, we obtain a classifier which is slightly less accurate but is still useful to be deployed in a real scenario. Second,
by deploying an adversarial training procedure,
we can reduce the attack surface and harden the resulting \gls{ids} while retaining all features and similar classification performance. 
We show that the \gls{ars} is significantly higher for the hardened model. 
\end{itemize}

We make all the source code, data, trained machine learning models and figures freely available to enable reproducibility and encourage experimentation at \url{https://github.com/CN-TU/adversarial-recurrent-ids}.

\section{An \gls{rnn}-based Classifier}


\begin{table}[b]
\caption{Flow occurrence frequency of attack types.}
\label{tab:occurrence}
\centering
\subfloat[CIC-IDS-2017\label{fig:cicids17proportions}
]{
\begin{tabular}{l r}
\toprule
Attack type & \hspace*{-4mm}Proportion \\ \midrule
DoS Hulk	&	10.10\%	\\
PortScan, Firewall	&	6.90\%	\\
DDoS LOIT	&	4.08\%	\\
Infiltration	&	3.30\%	\\
DoS GoldenEye	&	0.32\%	\\
DoS SlowHTTPTest	&	0.18\%	\\
DoS Slowloris	&	0.17\%	\\
Brute-force SSH	&	0.11\%	\\
Botnet ARES	&	0.03\%	\\
XSS attack	&	0.03\%	\\
PortScan, no Fw.	&	0.02\%	\\
Brute-force FTP	&	0.01\%	\\
SQL injection	&	$<$0.01\%	\\
Heartbleed	&	$<$0.01\%	\\
\bottomrule
\end{tabular}
}{}
\subfloat[UNSW-NB15\label{fig:unswnb15proportions}
]{
\begin{tabular}{l r}
\toprule
Attack type & \hspace*{-4mm}Proportion \\ \midrule
Exploits	&	1.42\%	\\
Fuzzers	&	1.01\%	\\
Reconnaissance	&	0.58\%	\\
Generic	&	0.21\%	\\
DoS	&	0.19\%	\\
Shellcode	&	0.08\%	\\
Analysis	&	0.03\%	\\
Backdoors	&	0.02\%	\\
Worms	&	0.01\%	\\
\bottomrule
\end{tabular}
}{}
\end{table}

We implemented a three-layer LSTM-based classifier with 512 neurons at each layer. For a large-enough network, we do not expect these architectural parameters to have a severe impact on classification accuracy, so we chose these parameters to obtain a good performance while keeping training duration at an acceptable level.

As the input features we use
source port, destination port, protocol identifier, packet length, \gls{iat} to the previous packet in the flow, packet direction (i.e. forward or reverse path) and all TCP flags (0 if the flow is not TCP).
We omitted \gls{ttl} values, as they are likely to lead to unwanted prediction behavior \cite{bachl_walling_2019}.  Among the used features, source port, destination port and protocol identifier are constant over the whole flow while the other features vary.
During flow extraction we used the usual 5-tuple flow key, which distinguishes flows based on the protocol they use and their source and destination port and IP address.
We use Z-score normalization to transform feature values to the range appropriate for neural network training. We ensured that our classifiers do not suffer from overfitting using a train/test split of 2:1.

For evaluation, we use the \textit{CIC-IDS-2017} \cite{sharafaldin_toward_2018} and \textit{UNSW-NB15} \cite{moustafa_unsw-nb15:_2015} datasets which each include more than 2 million flows of network data, containing both benign traffic and a large number of different attacks. Attacks contained in the datasets are shown in \autoref{tab:occurrence}.
\autoref{tab:performance_results} shows the achieved classification performance when evaluating metrics per packet and per flow and includes performance results for an MLP classifier from~\cite{bachl_walling_2019} for comparison.
As depicted, our \gls{rnn}-based classifiers achieve an accuracy that is similar to previous work based on these datasets~\cite{meghdouri_analysis_2018,bachl_walling_2019}. However, unlike these classifiers, our recurrent classifier has the advantage of being able to detect attacks already before the attack flows terminate.

\begin{table}
\caption{Performance metrics per packet and per flow. MLP values from \cite{bachl_walling_2019} are presented for comparison.} \label{tab:performance_results}
\centering
\begin{tabular}{l r r r r r r} \toprule
& \multicolumn{3}{c}{CIC-IDS-2017} & \multicolumn{3}{c}{UNSW-NB15} \\
	&	Packet	&	Flow	& MLP &	Packet	&	Flow & MLP	\\	\midrule
Accuracy	&	99.1\%	&	99.7\%	&	99.8\%	&	99.5\%	&	98.3\%	&	98.9\%	\\
Precision	&	97.0\%	&	99.7\%	&	99.9\%	&	83.4\%	&	78.6\%	&	84.5\%	\\
Recall	&	97.8\%	&	99.1\%	&	99.2\%	&	87.6\%	&	72.6\%	&	82.9\%	\\
F1	&	97.4\%	&	99.4\%	&	99.5\%	&	85.5\%	&	75.5\%	&	83.7\%	\\
Youden's J	&	97.2\%	&	99.0\%	&	99.1\%	&	87.3\%	&	71.9\%	&	82.3\%	\\
\bottomrule
\end{tabular}
\end{table}

\section{Adversarial Attacks}
\label{sec:adv}
We now investigate whether known \gls{aml} methods can be used for generating adversarial flows for our \gls{rnn}. \cite{hashemi_towards_2019} previously studied the behavior of several Network \glspl{ids} under adversarial attacks but unlike us did not investigate \glspl{rnn}.

A network packet contains significantly less features than, e.g., an image and, furthermore, most features such as TCP flags cannot be easily manipulated, as their manipulation might violate the protocol specifications and thus cause communication to fail. We identify the packet length and the \gls{iat} as features which are most likely to be exploited and thus choose them to be modified by the adversary. But even these features cannot be manipulated due to problem-specific constraints:
\begin{itemize}[leftmargin=0.5cm] 
\item Only packets can be manipulated which are transmitted by the attacker, except for botnet and backdoor traffic, which is entirely controlled by an attacker and thus only packets travelling in one direction can be manipulated.
\item Packets must not be smaller than initially, as otherwise less information could be transmitted.
\item \glspl{iat} must not decrease, as otherwise the physical speed of data transmission can be violated in some cases. An in-depth analysis of cases in which reduction of \glspl{iat} is legitimate is complex, so we generally disallowed \gls{iat} reductions.
\end{itemize}

Several \gls{aml} methods have been proposed in the recent literature, achieving different speed-quality tradeoffs:
\cite{szegedy_intriguing_2014} shows that it is possible to create adversarial samples for an image, which look similar to the original image but are classified wrongly. \cite{goodfellow_explaining_2015} develops the \gls{fgsm}, which makes easy generation of adversarial samples possible. \cite{papernot_crafting_2016} explores the use of \gls{aml} for \glspl{rnn}, but lacks important \gls{aml} methods. 
We implemented the following \gls{aml} methods.

\subsection{Carlini-Wagner}
We implemented the \gls{cw} \cite{carlini_towards_2017}, performing gradient descent on the optimization objective
\begin{equation} \label{eq:carliniWagner}
d(X,\tilde X) + \kappa  \max(Z(\tilde X), \delta).
\end{equation}
Here, $d(\cdot)$ is a distance metric and $\kappa \in \mathbb R^+$ is a parameter governing the tradeoff achieved between attack success and distance from the original flow. Furthermore, $Z(\cdot)$ denotes the neural network's logit output, $X$ denotes the original flow and $\tilde X$ the adversarial flow optimized by \gls{cw}.

$\delta \in \mathbb R$ is a parameter that determines how far an adversary wants to exceed the decision boundary.
In the original publication $\delta=0$, meaning that the network's decision for adversarial samples is just between attack and benign traffic. In the present context, we need to make sure that the classifier's prediction would actually be benign, even though a certain level of noise will be added to \glspl{iat} due to the network between attacker and victim. Hence, we introduced $\delta=-0.2$, corresponding to a prediction confidence of 55\% for the sample to be benign after the sigmoid activation function.

We used $L_1$ as distance metric, as we consider $L_1$ distance to represent practically relevant differences of network flows best. We used \gls{pgd} for meeting the real-world constraints discussed above.

\subsection{$L_\infty$-bounded Projected Gradient Descent}
\cite{madry_towards_2018} uses a method for generating adversarial samples which deploys \gls{pgd}  to maximize the network's loss function while constraining the achieved $L_\infty$ distance from the original samples. Hence, an important difference to \gls{cw} is that for this method the network's loss instead of its logit output is considered.
We consider this method an interesting combination of \gls{cw} and \gls{fgsm}.
Since its functioning is somewhat similar to \gls{cw}, we expected the generated adversarial samples to be similarly close to the original samples.

\subsection{Fast Gradient Sign Method}
Finally, we also tested the \gls{fgsm} 
 \cite{goodfellow_explaining_2015}, which is the first method for generating adversarial samples. FGSM can be considered a single pass of \gls{pgd} on the loss function with an equality constraint on the $L_\infty$ distance, i.e. the adversarial sample is found as
\begin{equation}
\tilde X = X + \epsilon \, \text{sgn}( \nabla_X L(X)),
\end{equation}
where $L(X)$ denotes the network's loss function and $\epsilon$ denotes the achieved $L_\infty$ distance.

\begin{figure}[h]
\includegraphics[width=\columnwidth]{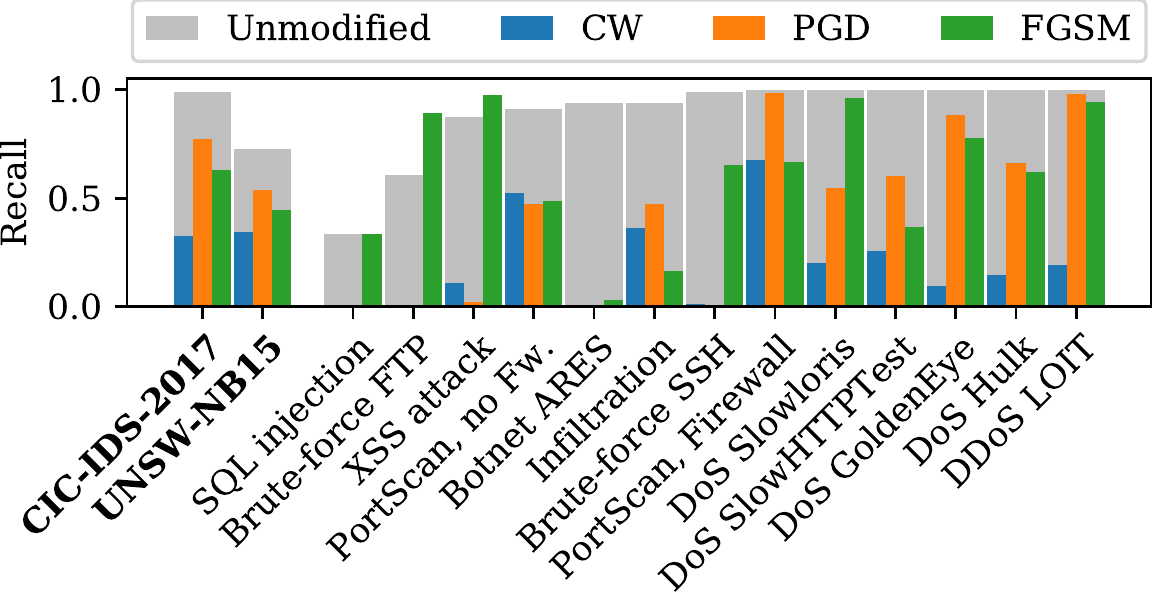}
\caption{Attack success ratios for both datasets and per attack type for CIC-IDS-2017. Depicted are flow detection accuracies for adversarial flows and for unmodified attack flows.}
\label{fig:adv_per_family}
\end{figure}

\subsection{Evaluation}
\autoref{fig:adv_per_family} depicts the performance for the investigated algorithms for both datasets and for each attack type in CIC-IDS-2017.
We used \gls{cw} with a tradeoff of 1 and, to provide a fair comparison, for \gls{fgsm} and \gls{pgd} we used an average $L_\infty$ distance as observed for \gls{cw}.

\gls{cw} delivers the best performance and FGSM, while being the fastest of all investigated algorithms, delivers the worst performance.
The figure shows significant differences for detecting different attack families in the first place. Also for generating adversarial flows some attack families are more susceptible than others. However, the results match to a high degree with our expectations. For example, SQL injection attacks apparently are closer to benign traffic than \gls{dos} attacks and, hence, finding adversarial samples should be easier.

Interestingly, any increase of the $L_\infty$ bound for \gls{pgd} did not yield significantly improved performance. 
\gls{cw} thus generally achieved superior results to $L_\infty$-bounded \gls{pgd}
and we can confirm the recommendation by \cite{carlini_towards_2017} to perform gradient descent on the logit output instead of the loss for good results.

\begin{figure}[b]
\includegraphics[width=\columnwidth]{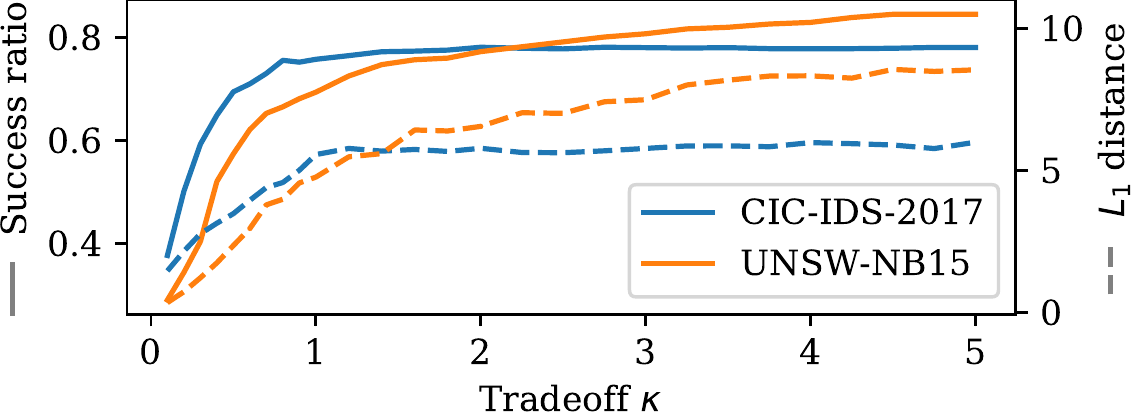}
\caption{Performance of \gls{cw} with different $\kappa$ values.}
\label{fig:adv_cw}
\end{figure}

\autoref{fig:adv_cw} depicts the success ratio and average distance for \gls{cw} for different tradeoffs $\kappa$.
Evidently, the larger $\kappa$, the better the attack works, but also distances from original flows become higher. With acceptable distances, we were able to generate adversarial samples for about 80\% of attack samples.

For successful adversarial samples, the second term in \autoref{eq:carliniWagner} becomes constant and, hence, is no longer relevant for optimization. Hence, $\kappa$ is irrelevant for the achieved distance, but governs if for one sample an adversarial sample is found.
However, when using a large $\kappa$ in \autoref{eq:carliniWagner}, to avoid overly large steps that impede convergence, it is necessary to reduce the gradient descent learning rate. Since the distance term is not affected by $\kappa$, it is furthermore necessary to scale the iteration count inversely with the learning rate. \gls{cw} with a large $\kappa$ therefore needs more time than with a small $\kappa$.


\section{Evaluating Adversarial Robustness}

As \autoref{fig:adv_per_family} shows, for most attack types, most samples can be modified by an adversary to be classified as benign by using \gls{cw}. High recall for a particular attack type does not imply that adversarial samples are hard to find  
for these attacks. Thus we argue that beside the classical metrics such as accuracy, false positives etc., a metric is needed, which quantifies how easily an \gls{ids} classifier can be fooled.

Such a metric should be easy to compute and easy to interpret.
While in general adversarial robustness for \gls{ml} models is frequently quantified as the average or median of minimum distances of adversarial samples \cite{carlini_evaluating_2019}, we found no generally agreed-upon metric for \glspl{ids} in the literature. If we consider flows, for which no adversarial sample can be found, to have infinite distance, the median has the advantage of ignoring such unsuccessful samples and outliers. On the other hand, the median might depict expected distances badly if they have a very uneven distribution. 

We thus propose the \gls{ars} as follows. Let $\mathcal S$ denote the set of samples, $N = | \mathcal S | \in \mathbb N$ the total number of samples and $d_s \in \mathbb R_0^+$ the distance of an adversarial flow from the original flow for a sample $s\in \mathcal S$, assigning unsuccessful samples a distance of $\infty$. 
We define the \gls{ars} as 
\begin{equation}
\text{ARS} = \frac{1}{\lceil N/2 \rceil} \sum\nolimits_{\tilde s \in \tilde {\mathcal S}} d_{\tilde s},
\end{equation}
with $\tilde {\mathcal S} \subset \mathcal S$ denoting a set of size $|\tilde {\mathcal S}| =\lceil N/2 \rceil$, so that $d_{\tilde s} \le d_s$ for all $\tilde s \in  \tilde {\mathcal S}, s \in   {\mathcal S} \setminus \tilde {\mathcal S} $.
The \gls{ars} is thus approximately the average of distances not larger than the median distance.

We consider an adversary successful if he can cause at least 50\% of all samples to be misclassified. In this case, the \gls{ars} is the average of the distances of the adversarial samples to the original samples for the 50\% of all samples which have the smallest distances. If the adversary doesn't manage to manipulate enough samples, the \gls{ars} is $\infty$.

\gls{cw} is able to find adversarial samples with minimum distance, but becomes slow for large $\kappa$. Hence, it can be used for finding the \gls{ars} efficiently as follows: 
Use \gls{cw} with a small tradeoff, trying to generate adversarial samples for an attack type. If 
 at least $N/2$  are wrongly classified, and
 the smallest distance of non-adversarial samples is not smaller than the $\lceil\frac{N}{2}\rceil$th smallest distance of adversarial samples,
stop and compute the \gls{ars}. Otherwise increase $\kappa$ and repeat. If after a predefined number of iterations -- e.g. 100 -- still not more than half of the samples are adversarial, also break and set the \gls{ars} to $\infty$.

The larger the \gls{ars} is, the more robust the model is, as then an adversary needs to modify the samples more to make them adversarial. If not more than half of the samples could be made adversarial, our metric is $\infty$ since then apparently it is not possible for an adversary to reliably conduct adversarial attacks on the majority of samples. In this case, the ratio of samples that could be made adversarial can be a useful metric to determine the exact extent of the adversarial threat.

Setting the threshold for the \gls{ars} to 50\% is arbitrary, but
 reasonable as outliers with very large distances are ignored and because if an attacker can make the majority of samples adversarial, we argue that its attack is ``successful''.

\begin{figure}[h]
\includegraphics[width=\columnwidth]{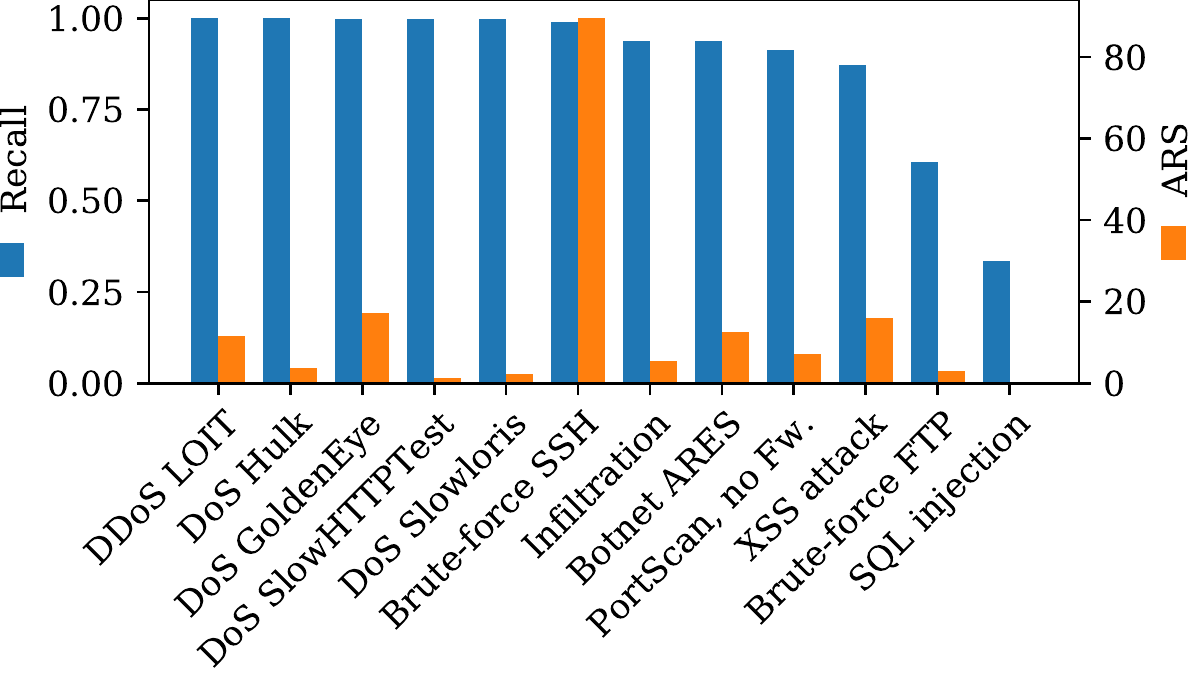}
\caption{Recall for unmodified flows and \gls{ars} for attacks in CIC-IDS-2017. For ``PortScan, Firewall'' recall never falls below 66\%: The attack does not succeed and the \gls{ars} is $\infty$.}
\label{fig:recall_ars}
\end{figure}

\autoref{fig:recall_ars} shows that certain attack types, like ``DoS SlowHTTPTest'', are easy to classify for an IDS but still are also surprisingly vulnerable to adversarial modifications by an attacker aiming to make them look benign.

\section{Explaining Recurrent Neural Networks}
Having verified the effectiveness of \gls{aml}  for our \glspl{rnn}, we now  investigate how the classifiers come to a decision.
From a naive perspective, one might be tempted to reuse existing explainability methods for \glspl{rnn} by considering a flow the sum of its packet features.
We identify several difficulties which occur when trying to explain decisions made by \glspl{rnn}.

\begin{itemize}[topsep=0pt,wide,labelwidth=!,labelindent=0pt]
\item
\textbf{Feature quantity.}
The number of features fed into an \gls{rnn} is the number of packet features times the length of the flow. For long flows, the total number of inputs can become very large.

\item
\textbf{Variable sequence lengths.}
The length of different flows might differ tremendously. Hence, features at one particular time step might be important for the network's outcome for one flow, but not even exist for other flows. 

\item
\textbf{Lack of a distance measure.}
However, even if we restricted the analysis to flows of a constant length, a flow is different from the plain concatenation of its packet features.
For example, in a sentence, which is sequence of words, parts can be rearranged, giving a different sequence with possibly the exact same meaning. 

\item
\textbf{Multiple prediction outputs.}
Often an \gls{rnn} produces an output at each time step. When  applying explainability methods, the question arises which output to consider for the method. The natural choice is to base the methods on the prediction output which occurs at the same time step as the feature under investigation: This approach is less complex compared to considering also features of all earlier time steps. Also, we expect the current prediction outcome to more dependent on the current feature, compared to a feature from many steps ago. 
However, due to the complex decision processes of deep neural networks,
this is not always true and a feature might influence a decision many time steps later. 
\end{itemize}

Many explainability methods proposed recently are local and thus provide explanations for a particular data sample \cite{shapley_value_1953,lundberg_unified_2017,dhurandhar_model_2018,ribeiro_why_2016}. However, for the particular problem of network traffic, due to the high number of flows and the characteristics of data, analyzing individual samples is of low interest. Explainability methods presented in this paper therefore aim to understand a model by analyzing which features are important, at which time step they are important and which feature values lead to classification as attack.


\subsection{Feature Importance Metrics}
\begin{table}
\caption{Accuracy drop for:}
\label{tab:feat_importance}
\centering
\subfloat[Input perturbation\label{fig:featRandom}
]{
\begin{tabular}{l r}
\toprule
Feature & \hspace*{-4mm}Accuracy~drop \\ \midrule
Protocol & 0.207 \\
Packet Length & 0.165 \\
SYN Flag & 0.099 \\
ACK Flag & 0.084 \\
Direction & 0.073 \\
Destination port & 0.071 \\
Source port & 0.060 \\
RST Flag & 0.057 \\
PSH Flag & 0.056 \\
Interarrival time & 0.024 \\
FIN Flag & 0.012 \\
URG Flag & 0 \\
ECE Flag & 0 \\
CWR Flag & 0 \\
NS Flag & 0 \\
\bottomrule
\end{tabular}
}{}
\subfloat[Feature dropout\label{fig:featDropout}
]{
\begin{tabular}{l r}
\toprule
Feature & \hspace*{-4mm}Accuracy~drop \\ \midrule
Destination port & 0.025 \\
Source port & 0.003 \\
RST Flag & 0.001 \\
ACK Flag & 0.001 \\
Protocol & 0.001 \\
Packet Length & 0.001 \\
Direction & 0.001 \\
SYN Flag & 0 \\
Interarrival time & 0 \\
FIN Flag & 0 \\
ECE Flag & 0 \\
URG Flag & 0 \\
CWR Flag & 0 \\
NS Flag & 0 \\
PSH Flag & 0 \\
\bottomrule
\end{tabular}
}{}
\end{table}
As a first step to understanding the neural network's decisions, we  estimate how important individual features are for the model's predictions.
When investigating an \gls{ml} classifier, it is natural to ask which inputs have a large influence on the classifier's prediction. We feel the need to distinguish metrics for this purpose based on their main aim:

A large amount of research has been spent on finding \emph{feature importance} metrics, which allow the selection  of high-importance features, providing a reasonably good classification performance while resulting in a more light-weight classifier.

Conversely, both adversarial machine learning and explainable machine learning bring up the question to what extent individual features are able to change the prediction outcome. While appearing similar, traditional feature importance can yield markedly wrong results for this case.  To distinguish, we propose the term \emph{feature sensitivity} for such metrics.
To analyze features, we use the following approaches:

\subsubsection{Neural Network Weights}
In previous works~\cite{olden_accurate_2004,olden_illuminating_2002}, a simple method for determining feature importance in neural networks has been summing up neural network weights leading from a certain input to the prediction outcome. The weights method can be considered for both feature importance and feature sensitivity, however, clearly, especially in the case of complex network architectures, this method is likely to provide wrong results. Hence, we provide results for the weights method mainly for comparison. Note that an LSTM cell alone has four weights leading from one input to an output.

\subsubsection{Input Perturbation}

The most commonly deployed feature importance techniques, used by practitioners for \glspl{rnn} \cite{stackexchange_cross_validated_neural_2019} and \gls{dl} \cite{molnar_interpretable_2019,stackexchange_cross_validated_feature_2016,olden_accurate_2004}, are based on adding noise to a particular feature and evaluating the drop in accuracy that occurs. We argue that it is hard to determine the ``correct'' intensity of noise to add. 
 Hence, we sample the value for a feature from the distribution of all values of this feature in the dataset. This makes the method non-parametric since the noise distribution  does not need to be chosen. We ensured that features which stay constant for a flow, i.e. source port, destination port and protocol, also stay constant throughout the flow when randomizing the feature.

\subsubsection{Feature Dropout}
While the perturbation method 
is convenient since it is easy to implement and understand, we argue that it has some shortcomings:
The \gls{rnn} was never trained for dealing with ``garbage'' values that the randomization creates. For example, completely unrealistic feature combinations could occur that were never observed during training. Furthermore, sequences of features could occur that cannot occur in reality.


To evaluate true feature importance, we thus develop a more sound method 
called \textit{feature dropout}: When training a model, for each sample, we leave out each feature  with independent probability $\frac{1}{n}$, $n \in \mathbb N$ being the number of features, by setting it to zero. On average, one feature gets zeroed out but it is also possible that none or more than one are left out. This procedure is equivalent to using dropout \cite{srivastava_dropout:_2014} with probability $\frac{1}{n}$ before the first layer.

An important implementation detail is that for each feature we add another input which is 1 if the feature is suppressed and 0 otherwise. This is necessary for the neural network to be able to distinguish between a feature missing and a feature genuinely being zero. The overall outcome is a classifier being able to deal with missing features.
The results in \autoref{tab:feat_importance} show that, unlike input perturbation, feature dropout  does not vastly overestimate features' importance. With feature dropout, it becomes apparent that only very few features  actually contain unique information, affecting accuracy when left out: the destination port and the source port.

A model trained with feature dropout typically yields slightly lower accuracy than a regularly trained model, even if no features are left out (flow accuracy of 99.43\% vs. 99.65\%). We thus recommend training two models: One regular one and one with feature dropout to use for the feature importance.

\begin{figure*}
\includegraphics[width=.49\textwidth]{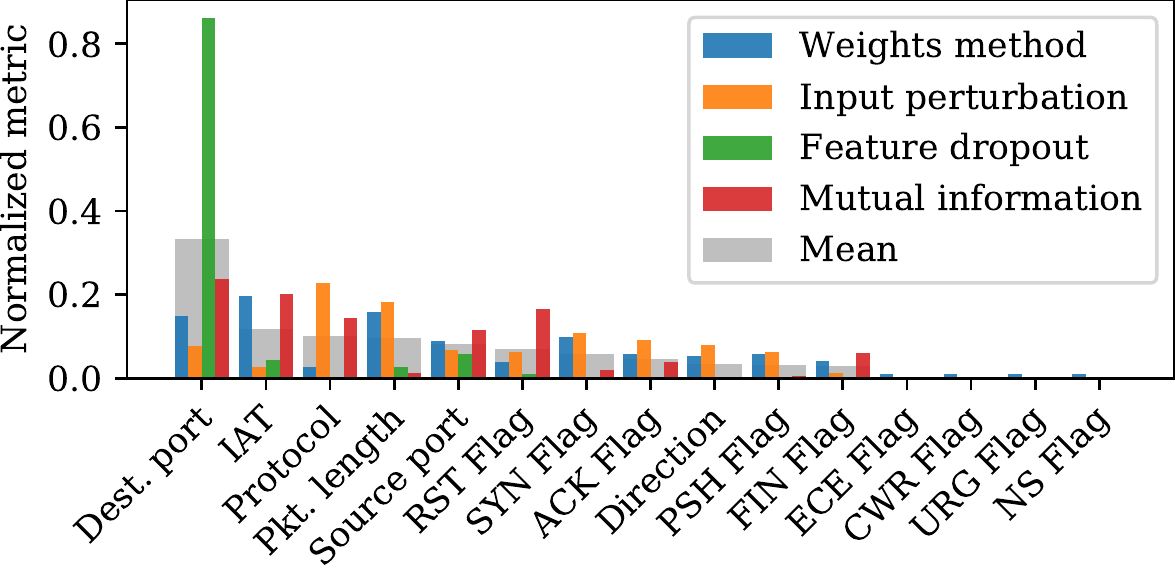}\hspace{.02\textwidth}
\includegraphics[width=.49\textwidth]{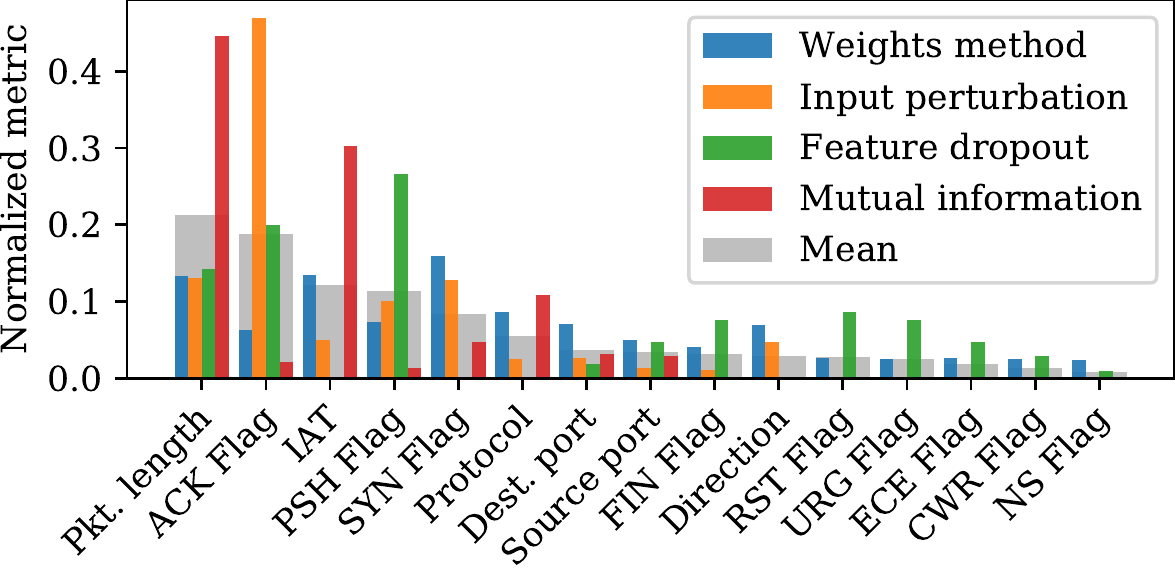}
\caption{Feature importance metrics for the flow prediction for CIC-IDS-2017 (left side) and UNSW-NB15 (right side).}
\label{fig:feat_imp}
\end{figure*}

\begin{figure}[b]
\includegraphics[width=\columnwidth]{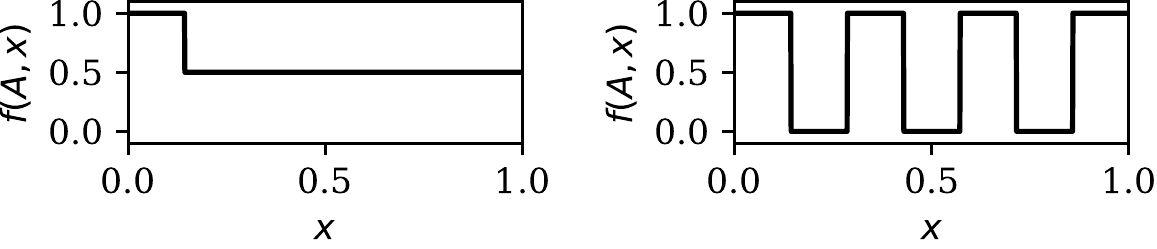}
\caption{Two distributions yielding an identical accuracy drop.}
\label{fig:mutinfo_example}
\end{figure}

Another method that uses dropout for feature importance is \cite{chang_dropout_2017}, applying a technique called \textit{Variational Dropout} to learn the optimal dropout rate for each feature. It tries to leave out as many features as possible and at the same time keep accuracy high. Thus important features are going to be left out less often and one can then extract the dropout probabilities for each feature and assess their significance based on them. %
While this method looks seemingly related to feature dropout, it is significantly more complex and does not aim to show the accuracy drop that occurs when omitting a feature but instead returns a unique feature importance value between 0 and 1. 

For feature dropout, it can also happen that several features are left out for a sample and so our method can also be used to analyze the effect of multiple features missing and can thus show possibly correlated features or -- more general -- features that contain common information, relevant for the classification task:
\begin{equation}
\text{score} = \frac{\text{acc}_{\text{base}}-\text{acc}_{\text{-both\_features}}}{\left(\text{acc}_{\text{base}}-\text{acc}_{\text{-feature\_1}}\right) + \left(\text{acc}_{\text{base}}-\text{acc}_{\text{-feature\_2}}\right)}
\end{equation}
With $\text{acc}_{\text{base}}$ we denote the accuracy of the classifier with all features included, with $\text{acc}_{\text{-feature\_i}}$ the accuracy if feature $i$ is omitted and with $\text{acc}_{\text{-both\_features}}$ the accuracy if both features are omitted. Assuming a non-negative accuracy drop when omitting a feature, the resulting score is $\ge 0.5$. The higher it is, the larger the information that both features share.

For example, the score between RST and the protocol identifier is 8.5; the highest of all pairs of features. While this may be unintuitive at the first glance, it likely stems from the fact that if the protocol identifier (TCP or UDP) is missing, then 
RST being 1 at some point indicates that the flow is TCP.

Feature dropout might constitute a building block for methods based on Shapley values \cite{shapley_value_1953} like KernelSHAP \cite{lundberg_unified_2017}. 
\subsubsection{Mutual Information}
Input perturbation and feature dropout mainly address feature importance.
For example, assuming that the test dataset is representative for production use, for feature importance it is reasonable to equate the distribution for perturbing a feature with the feature distribution itself. However, when evaluating feature sensitivity, e.g. for analyzing potential for adversarial samples, the attacker is not limited by this distribution and frequently is able to choose arbitrary values in the feature space.

Furthermore, we argue that accuracy drop depicts an importance measure which might be misleading for evaluating feature sensitivity.
To see this, let $f(A,x)$  denote the joint probability
for classification as attack and a feature value $x$. Accuracy then is $\int_{\mathbb R} f(A,x) dx$ for attacks and $1-\int_{\mathbb R} f(A,x) dx$ for benign traffic.
\autoref{fig:mutinfo_example} shows two different distributions
yielding the same accuracy, but clearly the right-side distribution has a larger influence on the prediction, as in the right-side case the prediction deterministically depends on the feature value.

To capture such dependencies, we propose to use mutual information 
to determine feature sensitivity.
Mutual Information between two random variables $X,Y$ is defined as
\begin{equation}
I_{X,Y} = \mathbb E \left\{ \log\left(\frac{f_{X,Y}(x,y)}{f_X(x)f_Y(y)}\right) \right\},
\end{equation}
with $f_X(x), f_Y(y)$ and $f_{X,Y}(x,y)$ denoting the distribution of $X,Y$ and their joint distribution, respectively. To obtain feature sensitivity, we compute mutual information between an input variable and the prediction output for one flow at one particular time step, averaging over the result for the test set.

\subsubsection{Comparison}
\autoref{fig:feat_imp} shows the results, which match to a large extent with domain-specific expectations for classifying network flows.
In particular, rarely used TCP flags like NS or URG are unimportant for the classifier. On the other hand, destination port and protocol are essential for characterizing flows by hinting at the type of traffic. \gls{iat} and packet length are important for estimating the amount of transferred information and several flags hint at certain attacks like \gls{dos} attacks.

The weights method is able to reveal features with a very low importance to a certain degree, but disagrees with the other methods to a large extent. Less anticipated, however, also input perturbation does not exhibit a considerable correlation with feature dropout. Considering its functioning of completely removing individual features, feature dropout is the most reliable method for evaluating importance for removing features. It is remarkable that none of the other methods is able to depict the distinct peak of importance for the destination port visible for feature dropout in \autoref{fig:feat_imp} and \autoref{tab:feat_importance}.

It is not surprising that mutual information disagrees with feature dropout, since both their aim and their functioning are substantially different. For example, mutual information shows that the protocol field can have a substantial impact on the classification even though an identical accuracy can be achieved when omitting it.


Metrics for UNSW-NB15 differ substantially from CIC-IDS-2017. However, due to the large number of different network attacks and network configurations, it is easily possible that relevant features are very different. We consider the question of model transferability of substantial importance for \gls{ids} applications, but out of scope for the present research.

\begin{figure}
\includegraphics[width=\columnwidth]{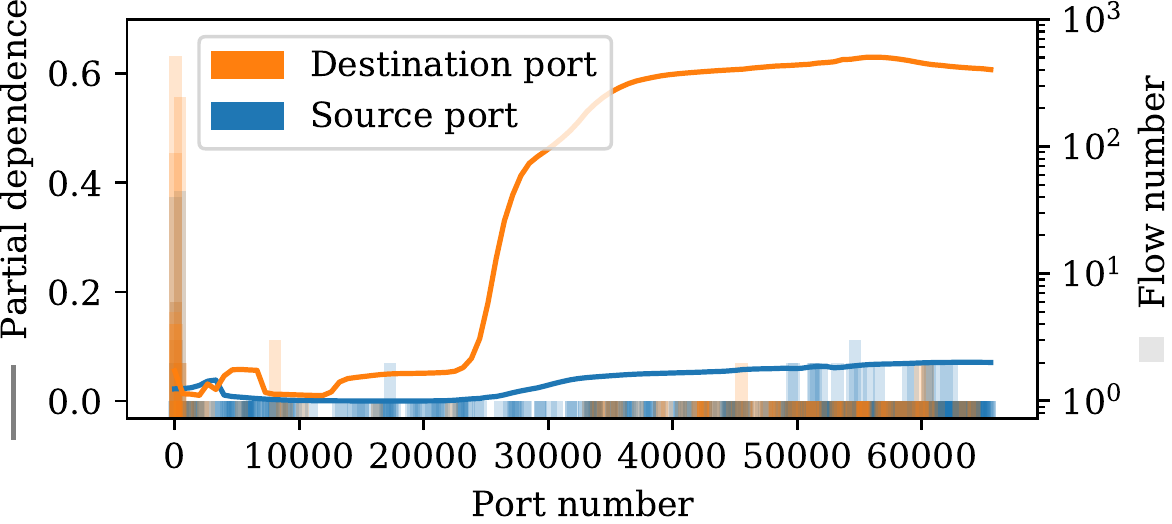}
\caption{PD plot for the source and destination port features.}
\label{fig:pdps}
\end{figure}

\subsection{Explainability Plots}
Knowing which features to investigate, it is important to analyze which feature values lead to classification as attack.
In literature, the use of Partial Dependency Plots (\gls{pdp}) has been proposed~\cite{friedman_greedy_2001}. To inspect attack types in detail, in this research we use a \textit{conditional} form of the \gls{pdp}. If $\boldsymbol X \in \mathbb R ^n$ denotes a random vector drawn from feature space, $f(\boldsymbol X) \in [0,1]$ the neural network's prediction and $c$ the attack type, we define the conditional \gls{pdp} for feature $i$  as
\begin{equation} \label{eq:pdp_conditional}
\text{\gls{pdp}}_{c,i}(w) = \mathbb E_{\boldsymbol X | C} \Big(f \left( X_1,\ldots,X_{i-1},w,X_{i+1},\ldots X_n \right) | c\Big),
\end{equation}
empirically approximating the conditional distribution by the observed subset that belongs to a certain attack type.

By using a classical \gls{pdp} we would likely lose information due to the averaging over very different types of traffic. However, for network traffic in particular, investigating each sample individually is not possible. Hence, the conditional \gls{pdp} provides the ideal compromise for our application.

Due to the use of a 5-tuple flow key, port numbers and the protocol identifier are constant for all packets in one flow.
Hence, we can consider the \gls{rnn} a regular classifier and reuse 
\glspl{pdp}, which have been proposed for non-recurrent classification methods, by plotting the \gls{rnn}'s flow prediction outcome over one of these  features.
The results show that for some attack types the port numbers play an important role. When looking at the \gls{pdp} for benign traffic samples in \autoref{fig:pdps}, 
it becomes apparent that traffic destined to a high destination port is generally indicative of an attack. We argue that this is because most services that regular users use have low port numbers.

\subsection{Plots for Sequences}

Intuitively, features at the beginning of a flow should be the most important while the classifier's predictions should not vary significantly anymore, as soon as it has come to a decision.
\begin{figure}
\includegraphics[width=\columnwidth]{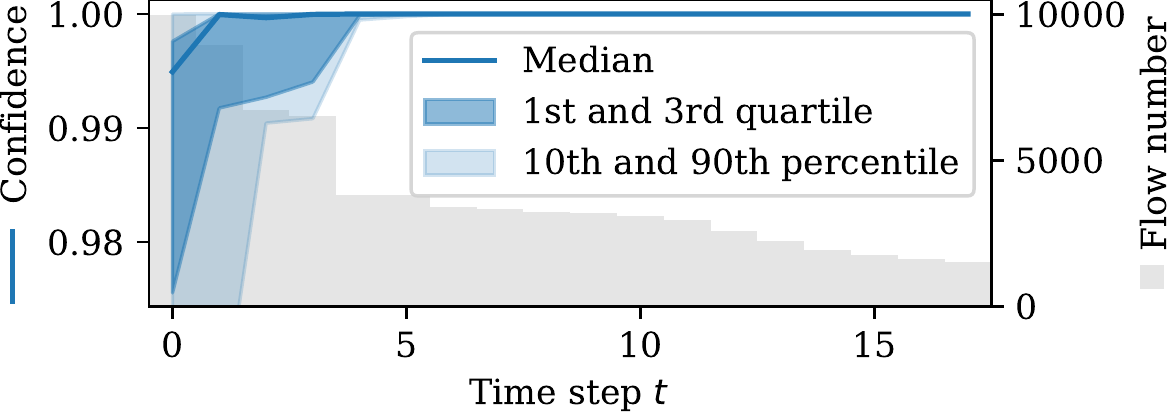}
\caption{Classifier confidence per time step for CIC-IDS-2017. 
For the majority of attack types, confidence increases in the first few steps and then stays almost constant at 1.}
\label{fig:confidence}
\end{figure}

To evaluate this hypothesis, \autoref{fig:confidence} depicts the classifier's prediction confidence for each time step, along with the number of samples having at least this length, which were used for evaluating the figure. 
While at the first couple of packets the confidence is not very high, towards the end of the flow it reaches values close to 1 and stays there. Hence, not only is the classifier able to make a reasonable classification after just a few packets, the figure also suggests that indeed later packets have a negligible influence on the prediction.

For investigating in more detail how features influence the prediction outcome, we extend \glspl{pdp} to the sequential setting.
Denoting as $X=\{\boldsymbol X^1, \ldots, \boldsymbol X^m\}$ the series of packet features $\boldsymbol X^t \in \mathbb R^n$ and $h_t(X)$ the network's hidden state after time step $t$, we define the sequential \gls{pdp} as
\begin{align}
\text{se}&\text{qPDP}_{c,i}(t,w)= \\ &\mathbb E_{X | C} \Big(f \left(h_{t-1}( X), X_1^t,\ldots,X_{i-1}^t,w,X_{i+1}^t,\ldots X_n^t \right) | c\Big). \nonumber
\end{align}

\begin{figure}[h]

\includegraphics[width=\columnwidth]{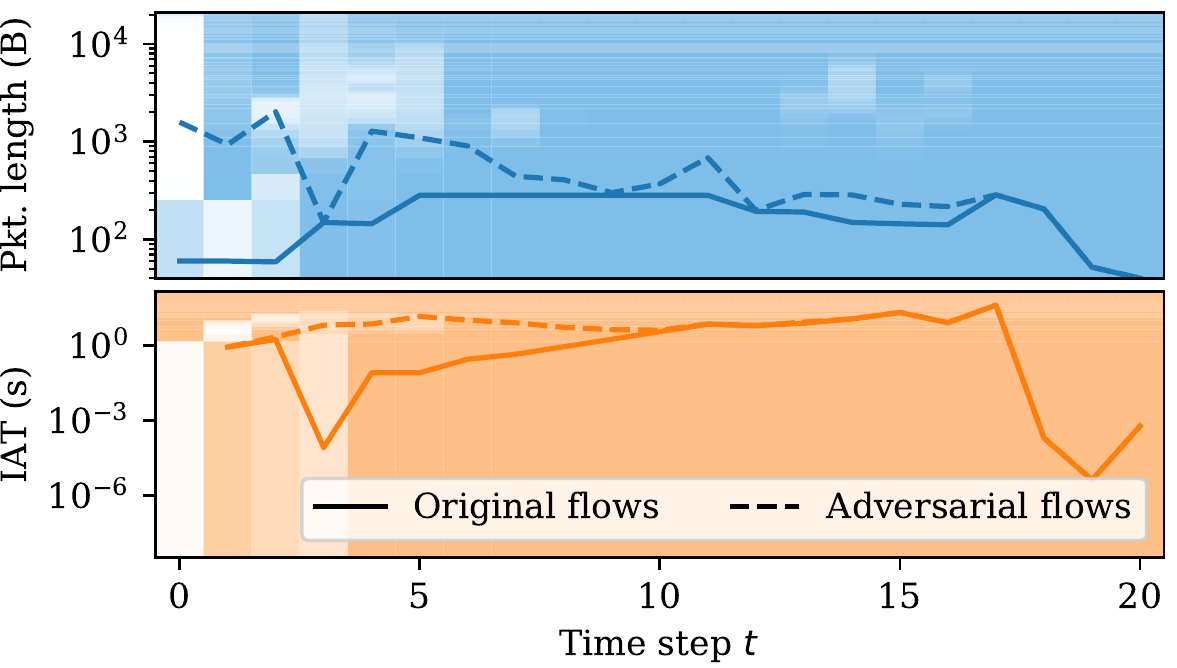}

\caption{Exemplary sequential PD plot and adversarial flows for the DoS Slowloris attack in CIC-IDS-2017. The lines show the feature's mean values. 
The shaded region shows the change in confidence that occurs when the feature is varied.}
\label{fig:pred_plots2}
\end{figure}

\autoref{fig:pred_plots2} shows an example together with the mean values for both unmodified samples and adversarial samples. Also in this figure we notice that mainly the first few packets are able to influence the prediction outcome and modifying features at a later time step does not change its confidence any longer. In many cases, the adversarial sample generation indeed moves packet features to areas where the network is less likely to be classified as attacks, confirming the effectiveness of \glspl{pdp}. In other cases, however, we did not observe an agreement between \gls{pdp} and adversarial modifications, hinting at dependencies which cannot be presented by \glspl{pdp}. Since the \gls{iat} is undefined for the first packet, we always set it to 0. Still, interestingly, the plot shows, that the classifier considers packets with a higher IAT to be more likely to be attacks than those with a smaller one.



\begin{figure}[t]
\includegraphics[width=\columnwidth]{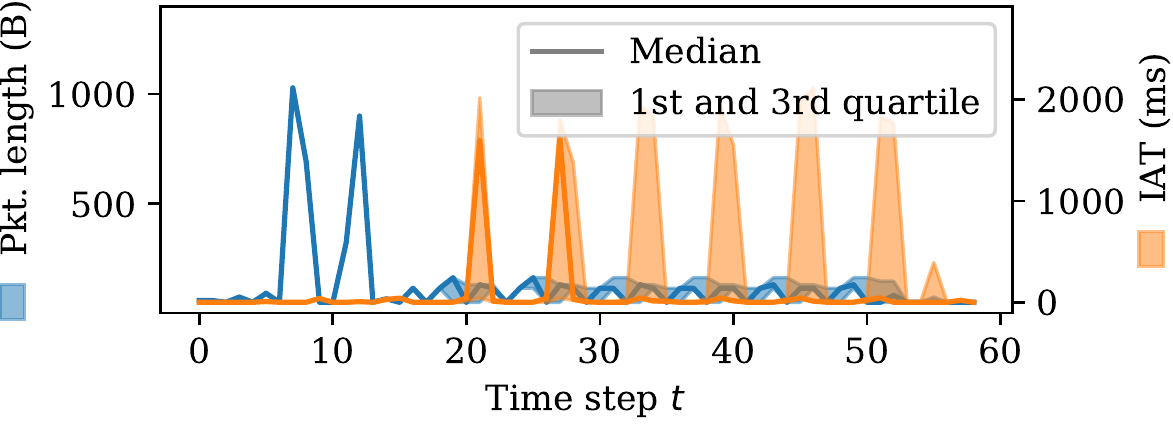}
\caption{\gls{iat} and packet length for SSH brute-force attacks in CIC-IDS-2017.}
\label{fig:plot_features}
\end{figure}

Finally, we investigated whether our classification task involves recognizing complex patterns in the feature space.
As example, \autoref{fig:plot_features} shows that
attack types indeed have a characteristic pattern in which they send packets by which they are easily recognizable.
Other attack families similarly show characteristic patterns.

\section{Defenses}
We now investigate two approaches to increase robustness of the \glspl{rnn} against adversarial attacks. Several methods have been proposed to improve robustness in the context of computer vision~\cite{akhtar_threat_2018}. We chose the methods presented below, because they are simple and can be applied to our recurrent setting in a straight-forward fashion.
\subsection{Reducing Features}
The most obvious defense is to simply omit features an attacker can manipulate. We try two different approaches of this defense strategy:
\begin{itemize}[leftmargin=0.5cm] 
\item Leaving out all manipulable features, i.e. packet size and \gls{iat}, in both directions.
\item Leaving them out only in the direction from the attacker to the victim. This, however, does not prevent adversarial samples for botnets, for which the attacker has control over both sides.
\end{itemize}

Both approaches lead to complete resistance to adversarial samples, except for botnets, which can still operate when only leaving out manipulable features in one direction. The results show that -- surprisingly -- there is only a small difference in classification performance when looking at flow accuracy: It is slightly lower at 99.3\% compared to 99.7\% originally for CIC-IDS-2017. However, packet accuracy is only 98\% when leaving out the features in one direction and 96.7\% when leaving them out in both directions. Thus, apparently the \gls{iat} and packet size are especially important for determining whether a flow is malicious in the first packets of a flow.

\subsection{Adversarial Training}
As alternative to omitting manipulable features, we attempted to make the classifier more robust against adversarial samples by augmenting the training set by adversarial flows generated using  \gls{cw}, labeled as additional attacks. This approach can be considered an adversarial training procedure, which is a common defense method in the related literature~\cite{akhtar_threat_2018,madry_towards_2018}.
We added one adversarial sample per attack sample to the training set. Since \gls{cw} is deterministic, this is the maximum number of adversarial samples possible. With this augmented training set we then alternated retraining of the neural network and optimization of the adversarial samples. 

A question which occurs in this process, is how many CPU cycles to spend on network training and adversarial sample optimization. We decided to use as many backpropagation steps for training as we use for adversarial sample optimization.  For each adversarial sample optimization step, we performed 10 iterations of gradient descent. Hence, all adversarial samples were optimized each 10 epochs of neural network training.

\begin{figure}[b]
\centering
\includegraphics[width=0.98\columnwidth]{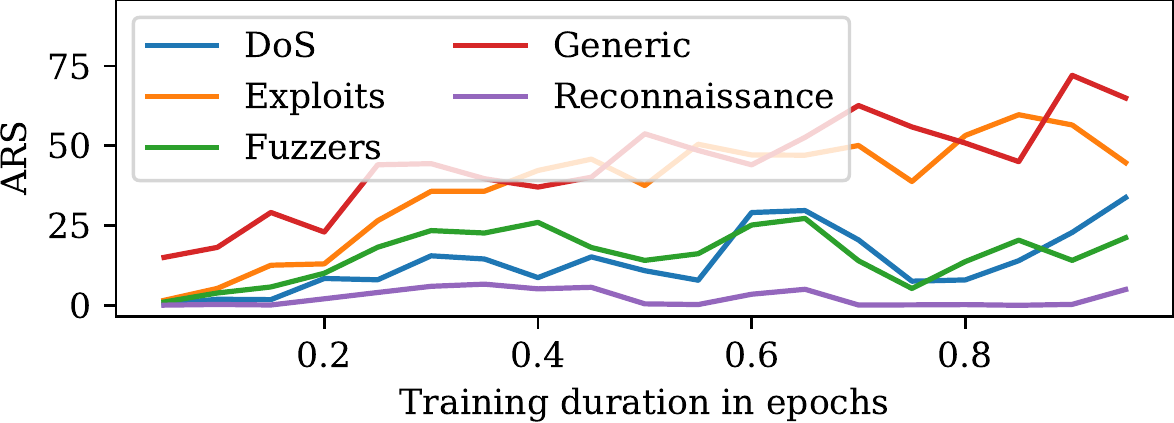}
\caption{\gls{ars} during adversarial training for UNSW-NB15.}
\label{fig:avdistance}
\end{figure}

\autoref{fig:avdistance} shows the \gls{ars} throughout  adversarial training  for several attack categories and clearly indicates that adversarial training is effective, as the distance gradually increases. Hence, an attacker would have to modify attack samples more and more, eventually rendering the attack unpractical.


For both datasets, after a small number of epochs, it was no longer possible to create a significant number of adversarial samples for most attack categories. 
After 50 epochs of training, accuracy is essentially identical to the results presented in \autoref{tab:performance_results}. Recall and informedness increased and precision slightly decreased. However, this is due to the higher proportion of attack samples in training data and, hence, expected.
We conclude that adversarial training is effective for reducing the attack surface for adversarial attacks in our scenario.

\section{Conclusion}

We have implemented a recurrent classifier based on LSTMs to detect network attacks, 
which is able to  detect attacks already before they are over. The recurrent approach allowed us to inspect the influence of single packets on the detection performance and shows which packets are \textit{characteristic} for attacks.
Even though the interpretation of \glspl{rnn} poses several difficulties, we have demonstrated methods for gaining insights into the model's functioning.

We showed that even though our use case is very different from computer vision, adversarial samples can be found efficiently, even if only ostensibly unimportant features can be modified. We introduced \textit{feature sensitivity} methods to show which features can easily be manipulated by an adversary to cause a wrong classification. We proposed the \gls{ars} for quantifying and comparing the adversarial threat for \glspl{ids}.
Deploying an adversarial training procedure, we could significantly reduce the adversarial threat.

\section*{Acknowledgements}
The Titan Xp used for this research was donated by the NVIDIA Corporation.

\renewcommand*{\bibfont}{\small}
\bibliographystyle{ieeetr}
\bibliography{bibliography}

\end{document}